\documentclass[conference,final,letterpaper,10pt]{IEEEtran}
\IEEEoverridecommandlockouts

\usepackage{textcomp}
\usepackage{xcolor}
\def\BibTeX{{\rm B\kern-.05em{\sc i\kern-.025em b}\kern-.08em
    T\kern-.1667em\lower.7ex\hbox{E}\kern-.125emX}}

\usepackage{cite}

\usepackage{graphicx}
\ifCLASSOPTIONcompsoc
    \usepackage[caption=false, font=normalsize, labelfont=sf, textfont=sf]{subfig}
\else
    \usepackage[caption=false, font=footnotesize]{subfig}
\fi

\usepackage{array}
\usepackage{textcomp}
\usepackage{stfloats}
\usepackage{url}
\usepackage{verbatim}

\usepackage{tabularx}
\usepackage{booktabs}
\usepackage{colortbl}

\usepackage{algorithm}
\usepackage{algpseudocode}

\usepackage{amssymb,amsmath,amsfonts,amsthm,bm,bbm,mathtools,cuted,bbold}

\usepackage[normalem]{ulem}

\usepackage[cmintegrals]{newtxmath}

\usepackage{multirow}

\DeclareMathOperator*{\argmax}{arg\,max}    

\newcommand{\ltwonorm}[1]{\left\lVert#1\right\rVert_2} 

\usepackage[acronym,,nohypertypes={acronym,notation}]{glossaries}
\newacronym{ccad}{CC-BAKD}{Communication-Constrained Bayesian Active Knowledge  Distillation}
\newacronym{rad}{RAKD}{Robust Active Knowledge Distillation}
\newacronym{akd}{AKD}{Active Knowledge Distillation}
\newacronym{bad}{BAKD}{Bayesian Active Knowledge Distillation}

\newacronym{kd}{KD}{Knowledge Distillation}

\newacronym{3d}{3D}{three dimensional}

\newacronym{al}{AL}{active learning}
\newacronym{arq}{ARQ}{automatic repeat request}

\newacronym{ial}{IAL}{Informative Active Learning}

\newacronym{cdma}{CDMA}{code division multiple access}
\newacronym{ccal}{CCAL}{Communication-Constrained Active Learning}
\newacronym{crc}{CRC}{cross-correlation}
\newacronym{ce}{CE}{cross-entropy}

\newacronym{rdft}{DCT}{Discrete Cosine Transform}

\newacronym{kl}{KL}{Kullbeck-Leiber}

\newacronym{mc}{MC}{Monte Carlo}
\newacronym{mkl}{MKL}{Kullbeck-Leiber}

\newacronym{elbo}{ELBO}{evidence lower bound}

\newacronym{onmse}{O-NMSE}{overall normalized mean-squared error}

\newacronym{nmse}{NMSE}{normalized mean-squared error}

\newacronym{oaal}{OAAL}{Over-the-Air Active Learning}

\newacronym{vi}{VI}{Variational Inference}

\newacronym{iqr}{IQR}{interquartile range}

\newacronym{dch}{DCH}{Data Channel}
\newacronym{fbch}{FBCH}{Feedback Channel}

\newacronym[plural=ACKs]{ack}{ACK}{acknowledgment}
\newacronym{aoa}{AoA}{angle of arrival}
\newacronym{awgn}{AWGN}{additive white Gaussian noise}
\newacronym{aod}{AoD}{angle of departure}

\newacronym{rat}{RAT}{reflected-angular training}
\newacronym[plural=APs, firstplural=access points (APs)]{ap}{AP}{access point}
\newacronym{b5g}{B5G}{Beyond-5G}
\newacronym[plural=BSs, firstplural=base stations (BSs)]{bs}{BS}{base station}

\newacronym{cc}{CC}{control channel}
\newacronym{chest}{CHEST}{channel estimation}
\newacronym{csi}{CSI}{channel state information}
\newacronym{cdf}{cdf}{cumulative distribution function}
\newacronym{crlb}{CRLB}{Cram\'er-Rao lower bound}

\newacronym{dc}{DC}{direct current}
\newacronym{dsp}{DSP}{digital signal processing}
\newacronym{dl}{DL}{downlink}
\newacronym{dlc}{DLC}{data link control}
\newacronym{doa}{DoA}{direction-of-arrival}
\newacronym{emf}{EMF}{electromagnetic field}
\newacronym{em}{EM}{electromagnetic}
\newacronym{fp}{FP}{fractional program}
\newacronym{glrt}{GLRT}{generalized likelihood ratio test}
\newacronym[plural=HRISs, firstplural=Hybrid Reconfigurable Intelligent Surfaces (HRISs)]{hris}{HRIS}{hybrid reconfigurable intelligent surface}
\newacronym[first=i.i.d.]{iid}{i.i.d.}{independent and identically distributed}
\newacronym{ios}{IoS}{Internet-of-Surfaces}
\newacronym{iot}{IoT}{Internet-of-Things}
\newacronym[plural=KPIs, firstplural=key performance indicators (KPIs)]{kpi}{KPI}{key performance indicator}
\newacronym{ls}{LS}{least-squares}
\newacronym{lf}{LF}{low frequency}
\newacronym{los}{LoS}{line-of-sight}

\newacronym{mac}{MAC}{medium access control}
\newacronym{mimo}{MIMO}{multiple-input multiple-output}
\newacronym{mmimo}{M-MIMO}{massive MIMO}
\newacronym{miso}{MISO}{multiple-input single-output}
\newacronym{ml}{ML}{machine learning}
\newacronym{mle}{ML}{maximum-likelihood estimator}
\newacronym{mmse}{MMSE}{minimum mean squared error}
\newacronym{mmtc}{mMTC}{massive machine-type communications}
\newacronym{mrc}{MRC}{maximum-ratio combining}
\newacronym{mse}{MSE}{mean-squared error}
\newacronym{nlos}{NLoS}{non-line-of-sight}

\newacronym{pca}{PCA}{principal component analysis}
\newacronym{pal}{PAL}{positive active learning}
\newacronym{phy}{PHY}{physical}
\newacronym[plural=PDFs]{pdf}{PDF}{probability distribution function}
\newacronym{pla}{PLA}{planar linear array}
\newacronym{pap}{P\&P}{plug-and-play}
\newacronym{ppp}{PPP}{Poisson point process}

\newacronym{ra}{RA}{random access}
\newacronym{rap}{RAP}{random access procedure}
\newacronym[plural=RISs, firstplural=reconfigurable intelligent surfaces (RISs), first=RIS]{ris}{RIS}{reconfigurable intelligent surface}
\newacronym{rf}{RF}{radio frequency}
\newacronym{rmse}{RMSE}{root-mean-square error}
\newacronym{rss}{RSS}{received signal strength}

\newacronym{saloha}{S-ALOHA}{slotted ALOHA}
\newacronym{se}{SE}{squared error}
\newacronym{sdp}{SDP}{semidefinite programming}
\newacronym{sdr}{SDR}{semidefinite relaxation}
\newacronym{ssl}{SSL}{self-supervised learning}
\newacronym{sic}{SIC}{successive interference cancellation}
\newacronym{sinr}{SINR}{signal-to-interference-plus-noise ratio}
\newacronym{smse}{SMSE}{sum mean squared error}
\newacronym{sdma}{SDMA}{space-division multiple-access}
\newacronym{snr}{SNR}{signal-to-noise ratio}
\newacronym{soa}{SoA}{state-of-the-art}
\newacronym{sre}{SRE}{smart radio environment}

\newacronym{skb}{SKB}{semantic knowledge base}

\newacronym{toa}{ToA}{time-of-arrival}

\newacronym{tdm}{TDM}{time-division multiplexing}
\newacronym{tdma}{TDMA}{time-division multiple access}
\newacronym{tdd}{TDD}{time-division duplex}
\newacronym{tem}{TEM}{transverse electromagnetic mode}

\newacronym[firstplural=users equipment (UEs), plural=UEs]{ue}{UE}{user equipment}
\newacronym{ul}{UL}{uplink}
\newacronym{ula}{ULA}{uniform linear array}
\newacronym{upa}{UPA}{uniform planar array}
\newacronym{uatf}{UatF}{use-and-then-forget}

\definecolor{amaranth}{rgb}{0.9, 0.17, 0.31}

\usepackage{lipsum}

\begin{document}

\title{
    {Batch Selection and Communication for Active Learning with Edge Labeling}
}

\author{
    {Victor~Croisfelt}, 
    {Shashi~Raj~Pandey}, 
    {Osvaldo~Simeone},
    and {Petar~Popovski} 
    \thanks{
        V. Croisfelt, S. R. Pandey, and P. Popovski are with the Connectivity Section of the Department of Electronic Systems, Aalborg University, Aalborg, Denmark (e-mail: \{vcr,srp,petarp\}@es.aau.dk). O. Simeone is with the King’s Communications, Learning \& Information Processing (KCLIP) lab within the Centre for Intelligent Information Processing Systems (CIIPS), Department of Engineering, King’s College London, London WC2R 2LS, U.K. (e-mail: osvaldo.simeone@kcl.ac.uk). O. Simeone is also a visiting professor in the Connectivity Section of the Department of Electronic Systems at Aalborg University. V. Croisfelt, S. R. Pandey, and P. Popovski were supported by the Villum Investigator Grant “WATER” from the Velux Foundation, Denmark, and the SNS JU project 6G-GOALS under the EU’s Horizon program Grant Agreement No 101139232. The European Union’s Horizon supported the work of O. Simeone and P. Popovski via the project CENTRIC (101096379). O. Simeone's work was also supported by an Open Fellowship of the EPSRC (EP/W024101/1), by the EPSRC project (EP/X011852/1), and by the UK Government via the FONRC Project REASON.
    }
}

\maketitle

\begin{abstract}
    Conventional retransmission (ARQ) protocols are designed with the goal of ensuring the correct reception of all the individual transmitter's packets at the receiver. When the transmitter is a learner communicating with a teacher, this goal is at odds with the actual aim of the learner, which is that of eliciting the most relevant label information from the teacher. Taking an active learning perspective, this paper addresses the following key protocol design questions: (\emph{i}) \emph{Active batch selection}: Which batch of inputs should be sent to the teacher to acquire the most useful information and thus reduce the number of required communication rounds? (\emph{ii}) \emph{Batch encoding}: Can batches of data points be combined to reduce the communication resources required at each communication round? Specifically, this work introduces \emph{Communication-Constrained Bayesian Active Knowledge Distillation} (CC-BAKD), a novel protocol that integrates Bayesian active learning with compression via a linear mix-up mechanism. Comparisons with existing active learning protocols demonstrate the advantages of the proposed approach. 
\end{abstract}

\begin{IEEEkeywords}
    Active knowledge distillation, Bayesian machine learning, compression.
\end{IEEEkeywords}

\section{Introduction}\label{sec:intro}

{I}{n} a classic communication-theoretic model~\cite{Bertsekas1996}, the sender organizes the data into a set of packets that are then passed on to a lower layer of the protocol stack. The responsibility of the lower layer is to ensure reliable transmission, such that \emph{all} data packets from the sender's set are reliably \textit{replicated} at the destination. In doing so, the sender runs an \gls{arq} protocol, and the destination sends ACK/NACK feedback messages to indicate the status of the received packets. The packets are transmitted without replacement; that is, upon reception of an ACK, the packet is never sent again. 

\begin{figure}[t!]
    \centering
    \includegraphics[width=\columnwidth, trim={1mm 6mm 0mm 5mm}]{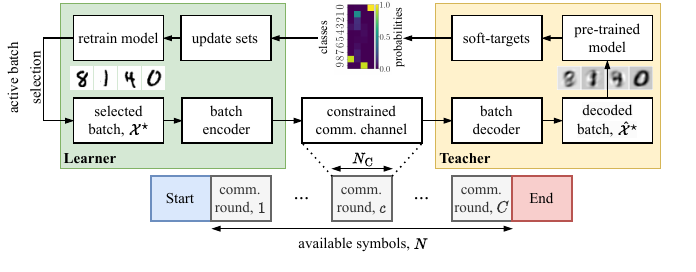}
    \caption{
        A learner communicates with a teacher over a constrained communication channel to obtain soft labels for batches of unlabeled inputs. This work aims to devise active batch selection strategies that use the available communication resources as efficiently as possible while reducing the communication cost of a batch through a batch encoding method.
    }
    \label{fig:system-setup}
\end{figure}

Now, assume that, as depicted in Fig.~\ref{fig:system-setup}, the sender is a learner that uses a channel to communicate with a teacher~\cite{Houlsb2011,Kirsch2019}. The packets at the learner encode unlabeled data samples that the learner can send over the channel to the teacher to obtain, possibly noisy, labels. This transmission could follow the traditional ARQ-based protocol, ensuring all $N$ data samples are replicated on the teacher side. This paper starts with the observation that the communication objective in this problem should  \textit{not} be to \textit{replicate}  data at the teacher, but rather to elicit the most informative label information from the teacher. 

This novel objective introduces two novel aspects in the design of the communication protocol. First, the learner can adaptively select the data points to communicate based on its current uncertainty about inference decisions at test time. Second, the learner may not need to encode data samples individually, requesting separate label information. Rather, a compressed mix-up over batches of selected inputs may suffice to obtain useful information from the teacher, saving bandwidth over the communication channel.


In this paper, we specifically  
address the following two design aspects for the setting in Fig. 1:

\noindent $\bullet$ \emph{Active batch selection}: Which inputs should be sent to the teacher to acquire the most useful information and thus reduce the number of required communication rounds?\\
\noindent $\bullet$ \emph{Batch encoding}: How do we encode the input information for transmission to the teacher to reduce the communication resources required at each round? 




\subsection{Active Batch Selection and Batch Encoding}
The problem of active batch selection can be viewed as a form of \emph{\gls{akd}}~\cite{Liang2022,Xu2023,Baykal2023}. In \gls{akd}, the goal is to select inputs for a teacher that responds with its model's predictive probabilities over the set of labels, \emph{i.e.}, with \emph{soft labels}, for the chosen inputs. Most existing \gls{akd} schemes select inputs with maximal predictive uncertainty for the learner's model~\cite{Xu2023,Baykal2023}. This approach suffers from \textit{confirmation bias}, as inputs with maximal predictive uncertainty for the learner may correspond to inherently hard examples to predict and thus to highly uncertain inputs for the teacher. The authors in~\cite{Baykal2023} proposed the \emph{\gls{rad}} protocol that strives to address the confirmation bias problem by making worst-case assumptions on the labeling errors made by the teacher, and it penalizes accordingly the potential information gains associated with each input. However, \gls{rad} does not account for communication constraints between learner and teacher.

Few papers have addressed the problem of batch encoding. In~\cite{Xu2023,Mishra2023}, the authors proposed to \emph{mix up} inputs within a batch, as in \cite{Zhang2018}, to augment the size of the unlabeled data set at the learner. The mix-up approach has not been explored in terms of its potential benefits in reducing communication requirements.

\subsection{Main Contributions}
In this paper, we contribute to both problems of active batch selection and batch encoding by proposing a communication protocol that integrates Bayesian active learning with compression via a linear mix-up mechanism. The approach, referred to as \emph{\gls{ccad}}, aims at maximizing the amount of reliable labeling information provided by the teacher per communication round.
\\\noindent $\bullet$ \emph{Bayesian \gls{akd} (BAKD)}: To reduce the number of communication rounds between the learner and the teacher, we adopt a Bayesian active learning strategy~\cite{Houlsb2011,Kirsch2019} that selects batches based on their \emph{epistemic uncertainty} at the learner's side. Epistemic uncertainty refers to the portion of the overall predictive uncertainty that additional data may decrease. As such, it contrasts with the inherent, aleatoric, predictive uncertainty that characterizes hard examples. By focusing solely on epistemic uncertainty, unlike \gls{rad}~\cite{Baykal2023}, the learner automatically addresses the confirmation bias problem of avoiding hard inputs characterized by significant inherent uncertainty.   
\\\noindent $\bullet$ \emph{Batch encoding via linear mix-up compression}: We propose a novel batch compression strategy that jointly compresses a batch of inputs over the feature dimension while mixing up the feature-compressed inputs. This compression strategy is integrated with the epistemic uncertainty-based active batch selection process to reduce the communication overhead per communication round. 

\glsunset{bad}
\noindent \emph{Organization:} The rest of the paper is organized as follows. In Section \ref{sec:setting}, we present the setting of interest. Section \ref{sec:bayesian-learning} presents a \gls{bad} protocol imposing only a maximum batch size constraint at each communication round. We then present the \gls{ccad} protocol in Section \ref{sec:ccal}. Experiments are given in Section \ref{sec:experiments}, followed by conclusions in Section \ref{sec:conclusions}.

\section{Setting}\label{sec:setting}
We study the setup illustrated in Fig.~\ref{fig:system-setup}, in which a learner aims to train a $K$-class classification model through communication with a teacher.

\subsection{Learning Model}
The learner has access to a \textit{pool set} $\mathcal{U}=\{\mathbf{x}_{i}\}_{i=1}^{U}$ with $U$ \textit{unlabeled} inputs, as well as an initial \textit{training set} $\mathcal{L}_0=\{(\mathbf{x}_{i},y_i)\}_{i=1}^{L}$ with $L$ \textit{labeled} examples, where $\mathbf{x}_{i}\in\mathbb{R}^{D}$ is the $i$-th input sample with $D$ features and $y_i\in\{0,\dots,K-1\}$ is the corresponding hard target. 

The goal of the learner is to train a classifier by leveraging a discriminative model $p(y|\mathbf{x},\boldsymbol{\theta})$, \emph{e.g.}, a neural network, with parameters $\boldsymbol{\theta}$. To this end, the learner wishes to \emph{distill} knowledge available at the teacher in the form of a \emph{pre-trained model} $p_{\rm tr}(y|\mathbf{x})$. For this purpose, at each \emph{communication round} $c\in\{1,\dots,C\}$, the learner selects a \textit{batch} of $B$ unlabeled inputs, denoted as $\mathcal{X}^{\star}=\{\mathbf{x}_1,\dots,\mathbf{x}_B\}$, from the pool set $\mathcal{X}^{\star}\subseteq\mathcal{U}$. The selected subset $\mathcal{X}^{\star}$ is removed from set $\mathcal{U}$ and sent to the teacher. The teacher responds to the learner with the \emph{soft targets} for each input from the batch  $\mathbf{x}_b \in \mathcal{X}^{\star}$, given by the $K\times{1}$ probability vector:
\begin{equation}
    p_{\rm tr}(\mathbf{x}_b) = [p_{\rm tr}(1|\mathbf{x}_b),\dots, p_{\rm tr}(K|\mathbf{x}_b)]^{\transp},
    \label{eq:true-soft-targets}
\end{equation}
which assigns a probability to each of the $K$ possible classes. 

After receiving feedback from the teacher, the learner updates the training set $\mathcal{L}$ with the corresponding soft targets. Thus, the updated training set $\mathcal{L}$ consists of two disjoint sets
\begin{equation}
    \mathcal{L}=\mathcal{L}_0\cup\mathcal{L}_{\rm tr},
    \label{eq:disjoint-sets}
\end{equation}
where the set $\mathcal{L}_0$ is the initial training set with hard targets $y_i$ and the set $\mathcal{L}_{\rm tr}$ denotes the part of the training set with the received soft targets $p_{\rm tr}(\mathbf{x})$ as in \eqref{eq:true-soft-targets}. Using the updated training set, the learner retrains its local model.

\subsection{{Communication Model}}\label{sec:communication-model}
At each communication round, the learner communicates with the teacher over a \emph{frame} of fixed length given by $N_{\rm C}$ real numbers~\cite{Bertsekas1996}. Letting $N$ be the total number of real numbers, also referred to as \emph{symbols}, available for communication from learner to teacher. Thus, the total number of communication rounds of the protocol is constrained and given by
\begin{equation}
    C=\left\lfloor{\frac{N}{N_{\rm C}}}\right\rfloor.
    \label{eq:number-of-steps}
\end{equation}
Each transmitted symbol is affected by additive i.i.d. zero-mean Gaussian noise with \textit{noise power} $\gamma^{-1}$.\footnote{Such noise models potential distortions caused by  \emph{quantization} \cite{zamir1996lattice} or by channel noise stemming from analog transmission (see, e.g.,~\cite{liu2022wireless}).} We consider the portion of the frame allocated for communication between teacher and learner to be negligible compared to that between learner and teacher. This assumption is reasonable because sending soft targets consumes significantly fewer resources than sending batches of input feature vectors; we do not model the communication resources from teacher to learner. 

\section{Bayesian Active Knowledge Distillation}\label{sec:bayesian-learning}
In this section, we present a \gls{bad} protocol that reduces the number of communication rounds while imposing only a maximum batch size constraint at each round. The following section will integrate this protocol with a batch compression strategy to yield the proposed \gls{ccad} algorithm.

\subsection{Bayesian Active Knowledge Distillation}\label{sec:active-distillation}
The goal of \gls{akd} is for the learner to reduce the number of communication rounds with the teacher~\cite{Baykal2023}. In this section, we propose to apply tools from Bayesian active learning to introduce a novel AKD protocol, referred to as \gls{bad}, that selects inputs with maximum epistemic uncertainty at the learner~\cite{Houlsb2011,Gal2016,Gal2017,Kirsch2019,Simeone2022}. As discussed in Section~\ref{sec:intro}, the underlying principle is that the learner should select a batch of inputs on which its current model has maximal epistemic uncertainty, thus avoiding inputs that are inherently hard to predict. 

\begin{algorithm}[h]
    \caption{\gls{bad} Protocol}
    \begin{algorithmic}[1]
        \State \textbf{Input:} initial training set, $\mathcal{L}=\mathcal{L}_0$; initial pool set, $\mathcal{U}$; batch size, $B$; number of communication rounds, $C$; initial distribution $q(\boldsymbol{\theta}|\mathcal{L})$; hyperparameter, $\tau$
        \For{step $c\in\{1,\dots,C\}$}
            \State learner selects and transmits a batch $\mathcal{X}^{\star}$ using~\eqref{eq:acquisition-step}
            \State get \textit{true soft targets} from teacher $\{p_{\rm tr}(\mathbf{x}^{\star}_b)\}_{\mathbf{x}^{\star}_b\in\mathcal{X}^{\star}}$
            \State $\mathcal{U}\gets\mathcal{U}\setminus\mathcal{X}^{\star}$
            \State $\mathcal{L}\gets\mathcal{L}\cup\{(\mathbf{x}_b^{\star},p_{\rm tr}(\mathbf{x}^{\star}_b))\}_{\mathbf{x}^{\star}_b\in\mathcal{X}^{\star}}$
            \State update distribution $q(\boldsymbol{\theta}|\mathcal{L})$ by minimizing~\eqref{eq:ad-free-energy}
            \If {$|\mathcal{U}|$ is 0}
                \State stop communication
            \EndIf
        \EndFor
    \end{algorithmic}
    \label{algo:bad}
\end{algorithm}

Accordingly, the learner decides a batch of inputs $\mathcal{X}^*\subseteq\mathcal{U}$ on which to query the teacher by using the \emph{BatchBALD acquisition function} in~\cite{Kirsch2019}. Specifically, the selected batch is obtained as
\begin{equation}
    \mathcal{X}^{\star}=\argmax_{\mathcal{X}\subseteq\mathcal{U}}
    a(\mathcal{X},q(\boldsymbol{\theta}|\mathcal{L})),
    \label{eq:acquisition-step}
\end{equation} 
where the BatchBALD acquisition function 
\begin{equation}
    a(\mathcal{X},q(\boldsymbol{\theta}|\mathcal{L}))=\mathbb{I}(y_1,\dots,y_B;\boldsymbol{\theta}|\mathcal{X},\mathcal{L})
    \label{eq:batchbald}
\end{equation} 
equals the mutual information between the labels $y_1,...,y_B$ corresponding to the selected inputs in the candidate set $\mathcal{X}=\{\mathbf{x}_1,...,\mathbf{x}_B\}$. The mutual information is evaluated by the learner concerning the joint distribution 
\begin{equation}
    p(\boldsymbol{\theta},y_1,...,y_B|\mathcal{X},\mathcal{L})=q(\boldsymbol{\theta}|\mathcal{L})\prod_{b=1}^B p(y_b|\mathbf{x}_b,\boldsymbol{\theta}),
\end{equation} 
where the \emph{variational posterior distribution} $q(\boldsymbol{\theta}|\mathcal{L})$ is maintained by the learner based on the current training set $\mathcal{L}$ as explained in the next subsection. 

The mutual information criterion in \eqref{eq:batchbald} captures the average \emph{disagreement} between the learner's models $p(y|\mathbf{x},\theta)$ over the batch of inputs when averaged with respect to distribution $q(\boldsymbol{\theta}|\mathcal{L})$. This average captures the level of epistemic uncertainty as predicted by the learner~\cite{Kirsch2019,Simeone2022}.

\subsection{Bayesian Learning with Soft Labels}
We now address the problem of optimizing the variational distribution $q(\boldsymbol{\theta}|\mathcal{L})$ based on the updated training set \eqref{eq:disjoint-sets}. To this end, we adopt the standard variational inference (VI) framework (see, e.g., \cite{Simeone2022}), which we extend to account for the availability of soft labels in $\mathcal{L}_{\rm tr}$. 

To start, let us define the standard \emph{free energy} criterion, also known as negative ELBO, for the labeled data set $\mathcal{L}_0$ as~\cite[p.456]{Simeone2022} 
\begin{align}
    L(q(\boldsymbol{\theta})|\mathcal{L}_0)=\mathrm{CE}(q(\boldsymbol{\theta})|\mathcal{L}_0) + \beta \cdot \mathrm{KL}(q(\boldsymbol{\theta})||p_0(\boldsymbol{\theta})) \text{ with } 
    \label{eq:free-energy}\\
    \mathrm{CE}(q(\boldsymbol{\theta})|\mathcal{L}_0)=\mathbb{E}_{(\mathbf{x},y)\sim\mathcal{L}_0}\left[\mathbb{E}_{\boldsymbol{\theta}\sim q(\boldsymbol{\theta})}\left[-\log p(y|\mathbf{x},\boldsymbol{\theta})\right]\right],
    \label{eq:ce}
\end{align} where (\ref{eq:ce}) represents the \gls{ce} loss, $p_0(\theta)$ is a fixed prior distribution, and the empirical average over the training set $\mathcal{L}_0$ is denoted as $\mathbb{E}_{(\mathbf{x},y)\sim\mathcal{L}_0}[\cdot]$. Furthermore, the constant $\beta>0$ is a hyperparameter that captures relative contributions of prior and data-dependent loss. 

Minimizing the free energy over the distribution $q(\theta)$ within some set of distributions yields the conventional VI algorithm. Instantiations of VI include \emph{variational dropout} (VD) \cite{Kingma2015,Gal2016,Molchanov2017}, where the variational distribution $q(\boldsymbol{\theta})$ is modeled as a Bernoulli-Gaussian vector. 

We extend VI to allow training over data set (\ref{eq:disjoint-sets}) consisting of both hard-labeled and soft-labeled examples. To this end, we propose to optimize the \emph{weighted free energy} criterion
\begin{equation}
    L(q(\boldsymbol{\theta})|\mathcal{L}_0,\mathcal{L}_{\rm tr}))=L(q(\boldsymbol{\theta})|\mathcal{L}_0) + \tau L(q(\boldsymbol{\theta})| \mathcal{L}_{\rm tr}),
    \label{eq:ad-free-energy}
\end{equation}
where the conventional free energy $L(q(\boldsymbol{\theta})|\mathcal{L}_0)$ is defined as in \eqref{eq:free-energy} as a function of the labeled data set $\mathcal{L}_0$, while the \emph{\textquotedblleft{soft}\textquotedblright free energy} criterion $L(q(\boldsymbol{\theta})|\mathcal{L}_{\rm tr})$ follows~\eqref{eq:free-energy} but with a modified \gls{ce} loss
\begin{equation}
    \mathrm{CE}(q(\boldsymbol{\theta})|\mathcal{L}_{\rm tr})=\mathbb{E}_{\mathbf{x}\sim\mathcal{L}_{\rm tr}}\left[\mathbb{E}_{\boldsymbol{\theta}\sim q(\boldsymbol{\theta})}\left[\mathbb{E}_{y\sim p_{\rm tr}({\mathbf{x}})}\left[-\log p(y|\mathbf{x},\boldsymbol{\theta})\right]\right]\right].
    \label{eq:ad-ce}
\end{equation}
The \gls{ce}~\eqref{eq:ad-ce} gauges the cross-entropy between the soft targets in~\eqref{eq:true-soft-targets} and the learner's predictive distribution. In the weighted free energy~\eqref{eq:ad-free-energy}, we introduce the hyperparameter $\tau\in\mathbb{R}_+$ to dictate the relative weight given by the learner to the feedback provided by the teacher as compared to the original labels in set $\mathcal{L}_0$. The \gls{bad} protocol is summarized in Algorithm \ref{algo:bad}.

\section{Communication-Constrained Bayesian Active Distillation}\label{sec:ccal}
In this section, we introduce \gls{ccad}, a generalization of the BAKD protocol introduced in the previous section that aims at reducing the required communication resources per communication round by compressing the selected batches at Step 3 of Algorithm~\ref{algo:bad}. However, compressing batches to reduce communication costs comes at the price of introducing \textit{reconstruction noise}, which other noise sources, such as \textit{quantization noise}, can further augment. Consequently, the batch received to be labeled by the teacher is distorted, which \emph{can} impair the quality of the output soft targets that the teacher can give as feedback to the learner. To combat this uncertainty, \gls{ccad} introduces a new acquisition function in contrast to the one in Step 3 of Algorithm~\ref{algo:bad}, as well as a novel model update in lieu of Step 7 of Algorithm~\ref{algo:bad}. An overview of \gls{ccad} is given in Algorithm~\ref{algo:ccad}. Details are given below.

\subsection{Batch Encoding}
After selecting a batch $\mathcal{X}^{\star}$ at a given step, as detailed in Section~\ref{sec:acquisition-function}, the learner compresses it for transmission. As we detail next, this is done using a linear compressor that implements a novel form of mix-up encoding.

Let $B^{\prime}\leq{B}$ and $D^{\prime}\leq{D}$ be the compressed batch and feature dimensions, respectively. Let us also denote as $\mathbf{X}^{\star}=[\mathbf{x}^{\star}_{1},\dots,\mathbf{x}^{\star}_{B}]\in\mathbb{R}^{D\times{B}}$ the data matrix for the selected batch and as $\mathbf{x}^{\star}=\mathrm{vec}\left(\mathbf{X}^{\star}\right)$ its vector form, where $\mathrm{vec}(\cdot)$ is the operation that stacks columns of the input matrix. We adopt a joint \emph{linear} compression scheme of the form
\begin{equation}
    \check{\mathbf{x}}^{\star}=\mathbf{Z}^{\transp}\mathbf{x}^{\star},
    \label{eq:encoder}
\end{equation}
where $\mathbf{Z}\in\mathbb{R}^{{BD}\times{B'D'}}$ is the \textit{compression matrix}. Note that the compressed vector $\check{\mathbf{x}}^{\star}$ in \eqref{eq:encoder} is of dimension $B^{\prime}D^{\prime}$. We use existing designs for matrix $\mathbf{Z}$. For example, matrix $\mathbf{Z}$ can be based on \gls{pca}~\cite{Bishop2006}. We further assume that the learner and the teacher agree on the compression matrix $\mathbf{Z}$ at the start of the protocol. 

\begin{algorithm}[h]
    \caption{\gls{ccad} Protocol}
    \begin{algorithmic}[1]
        \State \textbf{Input:} 
        initial training set, $\mathcal{L}=\mathcal{L}_0$; initial pool set, $\mathcal{U}$; batch size, $B$; number of communication rounds, $C$; initial distribution $q(\boldsymbol{\theta}|\mathcal{L})$; hyperparameter, $\tau$ compression matrix, $\mathbf{Z}$
        \For{step $c\in\{1,\dots,C\}$}
            \State \textbf{At learner:}
            \State select a batch $\mathcal{X}^{\star}$ using~\eqref{eq:new-acquistion-step}
            \State encode the selected batch using $\mathbf{Z}$ via \eqref{eq:encoder}
            \State \textbf{At teacher:}
            \State decode the compressed batch yielding $\hat{\mathcal{X}}^{\star}$ in~\eqref{eq:decoder}
            \State send \textit{estimated soft targets} $\{p_{\rm tr}(\hat{\mathbf{x}}_b^{\star})\}_{\hat{\mathbf{x}}_b^{\star}\in\hat{\mathcal{X}}^{\star}}$ in \eqref{eq:noisy-soft-targets}
            \State \textbf{At learner:}
            \State $\mathcal{U}\gets\mathcal{U}\setminus\mathcal{X}^{\star}$
            \State $\mathcal{L}\gets\mathcal{L}\cup\{(\mathbf{x}_b^{\star},p_{\rm tr}(\mathbf{x}^{\star}_b))\}_{\mathbf{x}^{\star}_b\in\mathcal{X}^{\star}}$
            \State update distribution $q(\boldsymbol{\theta}|\mathcal{L})$ by min.~\eqref{eq:ad-free-energy} w/~\eqref{eq:method1} or~\eqref{eq:method2}
            \If {$|\mathcal{U}|$ is 0}
                \State stop communication
            \EndIf
        \EndFor
    \end{algorithmic}
    \label{algo:ccad}
\end{algorithm}

Define the \emph{compression ratio} as
\begin{equation}
    R=\dfrac{BD-B^{\prime}D^{\prime}}{BD}.
\end{equation}
With a compression rate $R$, the number of communication rounds in \eqref{eq:number-of-steps} when $N_{\rm C}=B^{\prime}D^{\prime}$ evaluates to
\begin{equation}
    C=\left\lfloor{\dfrac{N}{B^{\prime}D^{\prime}}}\right\rfloor=\left\lfloor{\dfrac{N}{(1-R)BD}}\right\rfloor.
    \label{eq:new-al-steps}
\end{equation}
We use $R\in[0,1)$ to avoid the case in which communication is unconstrained: $C\xrightarrow[]{}\infty$ when $R=1$. The case where compression is not applied can be obtained by letting $R=0$, as is the case of \gls{bad}.

\subsection{Batch Decoding and Teacher's Feedback} 
The teacher receives the $B^{\prime}{D}^{\prime}$ vector
\begin{equation}
    \tilde{\mathbf{x}}^{\star}=\check{\mathbf{x}}^{\star}+\mathbf{n},
    \label{eq:received-signal}
\end{equation}
{where $\mathbf{n}\in\mathbb{R}^{B^{\prime}{D}^{\prime}\times 1}$ is the mentioned additive Gaussian noise accounting for quantization noise or analog communications, whose entries are distributed as $\mathcal{N}(0,\gamma^{-1})$ with mean zero and noise power $\gamma^{-1}$, as defined in Section~\ref{sec:communication-model}.} To estimate the selected batch, the teacher obtains the $D\times{B}$ vector
\begin{equation}
    \hat{\mathbf{x}}^{\star}=\mathbf{Z}\tilde{\mathbf{x}}=\mathbf{Z}\check{\mathbf{x}}^{\star} + \mathbf{Z}\mathbf{n}
    \stackrel{\text{(a)}}{=}\mathbf{Z}\mathbf{Z}^{\transp}\mathbf{x}^{\star}+ \mathbf{Z}\mathbf{n},
    \label{eq:decoder}
\end{equation}
where in (a) we used \eqref{eq:encoder}. 

We measure the \textit{overall distortion} due to transmission over the constrained communication channel as
\begin{equation}
    \ltwonorm{\hat{\mathbf{x}}^{\star}-\mathbf{x}^{\star}}^2=\ltwonorm{(\mathbf{Z}\mathbf{Z}^{\transp}-\mathbf{I})\mathbf{x}^{\star} + \mathbf{Z}\mathbf{n}}^{2},
    \label{eq:distortion}
\end{equation}
where, on the right-hand side, the first term within the norm represents the \textit{reconstruction noise} due to compression, and the second is the \textit{quantization or communication noise}. 

The decoded signal is then resized in matrix form as $\hat{\mathbf{X}}^{\star}=[\hat{\mathbf{x}}^{\star}_{1},\dots,\hat{\mathbf{x}}^{\star}_{B}]\in\mathbb{R}^{D\times{B}}$. The decoded batch of inputs is given by $\hat{\mathcal{X}^{\star}}=\{\hat{\mathbf{x}}^{\star}_1,\dots,\hat{\mathbf{x}}^{\star}_B\}$, where $\hat{\mathbf{x}}^{\star}_b$ is the $b$-th column of $\hat{\mathbf{X}}^{\star}$. Using~\eqref{eq:decoder}, the teacher labels the distorted outputs $\hat{\mathbf{x}}_b^{\star}\in\hat{\mathcal{X}}^{\star}$ to produce the \textit{estimated soft targets}
\begin{equation}
    p_{\rm t}(\hat{\mathbf{x}}_b^{\star}) = [p_{\rm t}(1|\hat{\mathbf{x}}_b^{\star}),\dots, p_{\rm t}(K|\hat{\mathbf{x}}_b^{\star})]^{\transp}.
    \label{eq:noisy-soft-targets}
\end{equation}
Note that the estimated soft targets in \eqref{eq:noisy-soft-targets} differ from the {true soft targets} $p_{\rm t}({\mathbf{x}^{\star}_b})$ in \eqref{eq:true-soft-targets} for $\mathbf{x}^{\star}_b\in\mathcal{X}^{\star}$ due to the noise in~\eqref{eq:distortion}.

\subsection{Learner's Model Update}
After receiving the teacher's feedback in \eqref{eq:noisy-soft-targets} for each input in the batch, the learner updates its training and pool sets, and it also updates its model to obtain a new distribution $q(\boldsymbol{\theta}|\mathcal{L})$, as per Steps 10-12 of Algorithm~\ref{algo:ccad}. As in \eqref{eq:disjoint-sets}, the updated training set is composed of two disjoint sets $\mathcal{L}=\mathcal{L}_0\cup\mathcal{L}_{\rm tr}$. However, unlike the setting studied in the previous section, set $\mathcal{L}_{\rm tr}$ is now impaired by the noise from~\eqref{eq:distortion} present in the estimated soft targets~\eqref{eq:noisy-soft-targets} provided by the teacher. To tackle this uncertainty, we propose two methods to the learner to obtain $q(\boldsymbol{\theta}|\mathcal{L})$, both of which generalize the weighted free energy criterion~\eqref{eq:ad-free-energy}.

\noindent \emph{Uncompressed covariates-based \gls{ce} loss:} The \gls{ce} loss is
\begin{equation}
\begin{aligned}
    &\mathrm{CE}_1(q(\boldsymbol{\theta}),\mathcal{L}_{\rm t})=\\
    &\mathbb{E}_{\mathbf{x}\sim\mathcal{L}_{\rm tr}}\left[\mathbb{E}_{\boldsymbol{\theta}\sim q(\boldsymbol{\theta})}\left[\mathbb{E}_{y\sim p_{\rm tr}(\hat{\mathbf{x}})}\left[-\log p(y|\mathbf{x},\boldsymbol{\theta})\right]\right]\right], 
    \label{eq:method1}
\end{aligned}    
\end{equation}

which associates the noiseless, clean inputs $\mathbf{x}$ to the estimated soft targets $p_{\rm tr}(\hat{\mathbf{x}})$. 

\noindent \emph{Compressed covariates-based \gls{ce} loss:} The \gls{ce} loss is
\begin{equation}
    \begin{aligned}
        &\mathrm{CE}_2(q(\boldsymbol{\theta}),\mathcal{L}_{\rm tr})=\\
        &\mathbb{E}_{\hat{\mathbf{x}}\sim\mathcal{L}_{\rm tr}}\left[\mathbb{E}_{\boldsymbol{\theta}\sim q(\boldsymbol{\theta})}\left[\mathbb{E}_{y\sim p_{\rm tr}(\hat{\mathbf{x}})}\left[-\log p(y|\mathbf{x},\boldsymbol{\theta})\right]\right]\right], 
        \label{eq:method2}
    \end{aligned}
\end{equation}
which associates the decoded, distorted inputs $\hat{\mathbf{x}}$ to the estimated soft targets $p_{\rm tr}(\hat{\mathbf{x}})$. Decoded inputs can be obtained at the learner's side by applying the encoder/decoder locally, but this method does not account for quantization noise. 

\subsection{Compression-Aware Active Batch Selection}\label{sec:acquisition-function}
In choosing the next batch $\mathcal{X}^{\star}$ from the pool $\mathcal{U}$, the learner should not only attempt to maximize the epistemic uncertainty at the current model, as in~\eqref{eq:acquisition-step} for BALD or BatchBALD~\cite{Houlsb2011,Kirsch2019}, but it should also account for the noise caused by the compression loss. For fixed compression matrix $\mathbf{Z}$, this can be done by letting the learner choose the batch on the decoded batch space $\hat{\mathcal{X}}=\{\hat{\mathbf{x}}_1,\dots,\hat{\mathbf{x}}_{B}\}$ using the encoder/decoder steps locally. Hence, we propose to generalize the acquisition function~\eqref{eq:acquisition-step} as
\begin{equation}
    {\mathcal{X}}^{\star}=\argmax_{\hat{\mathcal{X}}\in\mathcal{U}}
    a(\hat{\mathcal{X}},q(\boldsymbol{\theta}|\mathcal{L})),
    \label{eq:new-acquistion-step}
\end{equation}
where the decoded batch $\hat{\mathcal{X}}$ takes the place of the lossless covariates ${\mathcal{X}}$. We refer to the above generalization as the \textit{compression-aware acquisition function}. Note that this acquisition function does not deal with quantization noise.

\section{Experiments}\label{sec:experiments}
In this section, we empirically demonstrate the effectiveness of \gls{ccad} in Algorithm~\ref{algo:ccad} by comparing its performance with \gls{bad} in Algorithm~\ref{algo:bad}, which does not perform compression, and with \gls{rad} from \cite{Baykal2023}, which is based on an acquisition function that uses the aleatoric uncertainty at the learner. For clarity, when considering CC-BAKD, we show the best final average performance obtained with~\eqref{eq:method1} or~\eqref{eq:method2}. In this regard, we generally found that~\eqref{eq:method2} outperforms~\eqref{eq:method1} in situations of high compression ratio $R$. For \gls{bad} and \gls{ccad}, we display the best final average performance results regarding $\tau$ by performing a grid search over $\tau\in[0.001, 0.01, 0.1, 1, 10]$.

\subsection{Simulation Setup}\label{sec:experiments:setup}
Following ~\cite{Hinton2015,Gal2016,Gal2017}, we consider the MNIST dataset,  consisting of handwritten digit images with $D=28\times28=784$ pixels divided into $K=10$ digit classes. The standard MNIST training set of 60K examples is partitioned to create class-balanced sets $\mathcal{L}$ and $\mathcal{U}$ with $L=10$ and $U=1$K inputs, respectively. We use the remaining examples to create class-balanced training and validation sets for the teacher, with the validation set having a size of 100 examples and the training set having a size of 50K examples. This process is repeated ten times for statistical evaluation. The standard test set of 10K examples is used to evaluate the models of the learner and the teacher. 

\noindent \emph{Model architectures}: The learner has a neural network with two hidden layers with 800 ReLU units per layer. As in \cite{Gal2017}, VD is applied only in the last layer with a dropout probability of 0.5. The teacher has a neural network with two hidden layers with 1200 ReLU units per layer and three dropout layers with hidden dropout probabilities of 0.5 and an input dropout probability of 0.8. For VD, we used 10-100 stochastic realizations and Bernoulli dropout layers~\cite{Kingma2015,Gal2016,Molchanov2017}.

\noindent \emph{Training}: We use stochastic gradient descent with a training batch size of 32, a learning rate of 0.01, and a momentum of 0.9 over ten epochs for the learner's training. We further assume that the learner's training data $\mathcal{L}$ keeps the ordering given by the acquisition step. For the teacher's training, we use a learning rate of 0.01, a momentum of 0.9, and a weight decay of 5e-4. We employ early stopping during teacher training with a patience of 5 epochs. {The baseline test accuracy performance of the learner prior to communications is $31.8\pm5.8$ [\%], while the accuracy of the teacher's model is $96.0\pm0.6$ [\%].}

\noindent\emph{Communication model}: Unless otherwise stated, we consider a {worst-case scenario} where the number of available symbols $N$ is equal to the feature dimension $N=D=784$ symbols, meaning that if there is no compression, the number of communication rounds in~\eqref{eq:number-of-steps} would evaluate to one; that is, the learner would be able to query the label of a single input.

\noindent \emph{RAKD Benchmark}: 
To benchmark the performance of \gls{bad} and \gls{ccad}, we consider the \emph{Oracle} \gls{rad} method, which provides an upper bound on the accuracy achievable by RAKD~\cite{Baykal2023}. 



\noindent \emph{Linear compression}: We adopt \gls{pca} to obtain the overall compression matrix $\mathbf{Z}$ in \eqref{eq:encoder} by using a balanced 10K labeled examples dataset. A sub-sampling approach is used to generate data when $B>1$. This method creates random batches by selecting inputs from the 10K balanced without replacement. We repeat this process ten times, as before. 
Due to the limited variability of the MNIST dataset, we found that the region of interest for the compression ratio $R$ lies in $R\in[0.95,0.999]$. This is because most of the variability of the MNIST digits can be explained with few feature dimensions, and reconstruction noise is usually extremely low outside this range. 

\begin{figure}[t]
    \centering
    \includegraphics[trim=2mm 2mm 0 0, clip]{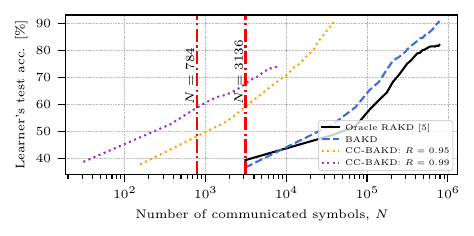}
    \caption{Evolution of the learner's performance as a function of the number of communicated symbols $N$. The batch size is $B=4$ for all schemes. The red lines indicate values of the number of symbols $N$ 
    required to transmit a single uncompressed input and a batch of $B=4$  uncompressed inputs, respectively.}
    \label{fig:comm}
\end{figure}
%

\subsection{Results without Quantization Noise}
\begin{figure}[t]
    \centering
    \includegraphics[trim=2mm 2mm 0 0, clip]{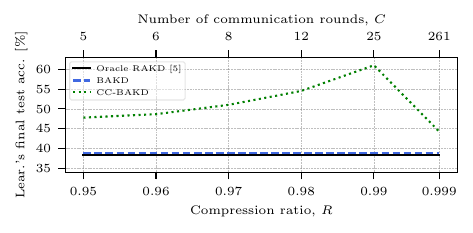}
    \caption{
        {Learner's final test accuracy as a function of the compression ratio, $R$, for a total number of symbols of $N=784$ and a batch size of $B=4$. For~\gls{ccad}, we show the corresponding number of communication rounds, $C$ in ~\eqref{eq:new-al-steps} in the top horizontal axis. For \gls{bad} and \gls{rad}, the number of communication rounds is fixed to one for a batch size of one since these protocols do not apply compression.}
    }
    \label{fig:compression-factor}
\end{figure}

We analyze the reconstruction noise in isolation by setting the additive noise power $\gamma^{-1}$ in~\eqref{eq:received-signal} to zero. We start by evaluating the impact of the compression ratio, $R$, for a fixed total number of symbols $N=784$ and a fixed batch size of $B=4$. Fig.~\ref{fig:compression-factor} shows the learner's final test accuracy after communicating with the teacher.  As $R$ increases, the performance of the benchmark methods \gls{bad} and Oracle \gls{rad} does not change since they do not apply compression. In contrast, for \gls{ccad}, as $R$ increases, more communication rounds are possible, improving the accuracy to some level, after which additional compression deteriorates the learner's performance. 

Figure~\ref{fig:comm} shows the evolution of the accuracy of different schemes as a function of the number $N$ of communicated symbols, which reflects the transmission latency. Take the point $N=784$ as an example. For this value of $N$, \gls{ccad}, with $B=4$ and $R=0.99$, attains an accuracy of approximately $61$\% while, with $B=4$ and $R=0.95$, the accuracy at $N=784$ is approximately $48$\%. In contrast, \gls{bad} and \gls{rad} cannot transmit any input for values of $N$ smaller than $N=784$. In this regard, \gls{bad} attains an accuracy of $60$\% only after approximately $65\times10^4$ symbols, implying that, to achieve the same accuracy of CC-BAKD, \gls{bad} requires a latency that is approximately 85 times larger than CC-BAKD.

\subsection{Results with Quantization Noise}
\begin{figure}[!htbp]
    \centering
    \includegraphics[trim=2mm 2mm 0 0, clip]{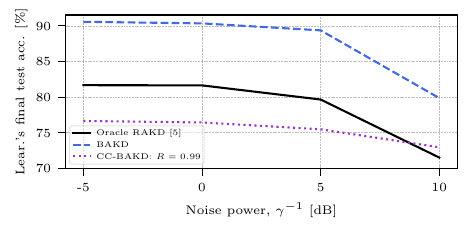}
    \caption{
    \gls{ccad} final learner's performance as a function of the noise power 
    for a batch size of $B=4$. For \gls{ccad} with $R=0.99$, the number of transmitted symbols is $N=7840$, while for \gls{rad} and \gls{bad}, the number of transmitted symbols is $N=78400$. 
    }
    \label{fig:quantization}
\end{figure}

We now study the impact of additive noise, which 
further degrades the estimated inputs at the teacher and, as a result, also the soft targets provided by the teacher to the learner. Figure~\ref{fig:quantization} shows the learner's final test accuracy as a function of the noise power $\gamma^{-1}$ in~\eqref{eq:received-signal} for a fixed batch size $B=4$. In this figure, it is noteworthy that \gls{bad} and \gls{rad} exhibit superior performance compared to \gls{ccad} because they utilize more symbols ($10 \textrm{ times}$). While noise generally degrades the learner's accuracy, \gls{ccad} is seen to be significantly more robust to the presence of noise than \gls{bad} and \gls{rad}. {This is because \gls{bad} and \gls{rad} curves decay faster as the noise power increases than the \gls{ccad} one.}

%

\section{Conclusions}\label{sec:conclusions}
This paper has introduced Communication-Constrained Bayesian Active Distillation (CC-BAKD), a new protocol that moves away from the classical ARQ-based approach of ensuring the correct reception of all the information packets at the transmitter, aiming instead to collect the most relevant information from the remote teacher. To do so, CC-BAKD  builds on Bayesian active learning to address the problem of confirmation bias and compression based on a linear mix-up mechanism. Numerical results have demonstrated clear advantages over the state-of-the-art. Future work may consider extensions involving multiple learners and/or teachers. 


\bibliographystyle{IEEEtran}

\end{document}